\documentclass[final]{cvpr}

\usepackage{times}
\usepackage{epsfig}
\usepackage{graphicx}
\usepackage{amsmath}
\usepackage{amssymb}

\usepackage{comment}
\usepackage{color}
\usepackage[font=footnotesize,skip=1pt,belowskip=-10pt]{caption}
\usepackage{subcaption}
\usepackage{multirow}
\usepackage{pdflscape}
\usepackage{tabularx,booktabs}
\usepackage{microtype}
\usepackage{enumitem}
\usepackage{array}
\usepackage{siunitx}

\usepackage{adjustbox}

\newcolumntype{R}[2]{%
    >{\adjustbox{angle=#1,lap=\width-(#2)}\bgroup}%
    l%
    <{\egroup}%
}
\newcommand*\rot{\multicolumn{1}{R{90}{1em}}}% no optional argument here, please!

\newcommand{\smallsec}[1]{\vspace{0.03in} \noindent {\bf #1:}}

\definecolor{amethyst}{rgb}{0.6, 0.4, 0.8}

\newcommand{\wa}{\mathbf{w_t}}
\newcommand{\wg}{\mathbf{w_g}}
\newcommand{\wat}{\wa^T}
\newcommand{\wgt}{\wg^T}
\newcommand{\z}{\mathbf{z}}

\usepackage[pagebackref=true,breaklinks=true,colorlinks,bookmarks=false]{hyperref}

\def\cvprPaperID{1263} % *** Enter the CVPR Paper ID here
\def\confYear{CVPR 2021}

\begin{document}

%%%%%%%%% TITLE
\title{Fair Attribute Classification through Latent Space De-biasing}

\author{Vikram V. Ramaswamy, Sunnie S. Y. Kim, Olga Russakovsky \\
Princeton University \\
{\tt\small \{vr23, suhk, olgarus\}@cs.princeton.edu}
}

\maketitle

%%%%%%%%% ABSTRACT (4000 characters max)
\begin{abstract}
   Fairness in visual recognition is becoming a prominent and critical topic of discussion as recognition systems are deployed at scale in the real world. Models trained from data in which target labels are correlated with protected attributes (e.g., gender, race) are known to learn and exploit those correlations. In this work, we introduce a method for training accurate target classifiers while mitigating biases that stem from these correlations. 
   We use GANs to generate realistic-looking images, and perturb these images in the underlying latent space to generate training data that is balanced for each protected attribute. We augment the original dataset with this generated data, and empirically demonstrate that target classifiers trained on the augmented dataset exhibit a number of both quantitative and qualitative benefits. We conduct a thorough evaluation across multiple target labels and protected attributes in the CelebA dataset, and provide an in-depth analysis and comparison to existing literature in the space. Code can be found at \url{https://github.com/princetonvisualai/gan-debiasing}. %\{add link to code}
\end{abstract}

%%%%%%%%% BODY TEXT
\section{Introduction}

\begin{figure}
    \centering
    \includegraphics[width=\linewidth]{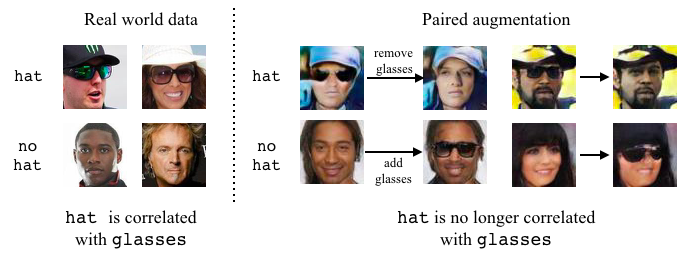}
    \caption{Training a visual classifier for an attribute (e.g., \texttt{hat}) can be complicated by correlations in the training data. For example, the presence of hats can be correlated with the presence of glasses. We propose a dataset augmentation strategy using Generative Adversarial Networks (GANs) that successfully removes this correlation by adding or removing glasses from existing images, creating a balanced dataset.
    }
    \label{fig:pullfig}
\end{figure}

Large-scale supervised learning has been the driving force behind advances in visual recognition. Recently, however, there has been a growing number of concerns about the disparate impact of these visual recognition systems. Face recognition systems trained from datasets with an underrepresentation of certain racial groups have exhibited lower accuracy for those groups~\cite{BG18IntersectionalDataset}. Activity recognition models trained on datasets with high correlations between the activity and the gender expression of the depicted person have over-amplified those correlations~\cite{ZWYOC17BiasAmp}. Computer vision systems are statistical models that are trained to maximize accuracy on the majority of examples, and they do so by exploiting the most discriminative cues in a dataset, potentially learning spurious correlations.
In this work, we introduce a new framework for training computer vision models that aims to mitigate such concerns, illustrated in Figure~\ref{fig:pullfig}.

One proposed path for building `fairer' computer vision systems is through a `fairer' data collection process. Works such as \cite{BG18IntersectionalDataset,YQFDR20BalancingImagenet} propose techniques for better sampling data to more accurately represent all people. Creating a perfectly balanced dataset, however, is infeasible in many cases.
With the advances in Generative Adversarial Networks (GANs)~\cite{GPMXWOCB14GANs}, several works propose using generated data to augment real-world datasets~\cite{GCSE19FairModeling,SHCV19FairnessGAN,XYZW18Fairgan}. These methods have been growing in computational and algorithmic complexity (e.g., \cite{SHCV19FairnessGAN,XYZW18Fairgan} adding multiple loss functions to GAN training), necessitating access to a sufficient number of inter-sectional real-world samples. In contrast, we demonstrate a simple and novel data augmentation technique that uses a single GAN trained on a biased real-world dataset.

\smallsec{Illustrative example} 
Consider our example from Figure~\ref{fig:pullfig}. Our goal is to train a visual recognition model that recognizes the presence of an attribute, such as wearing a hat. Suppose in the real world wearing a hat is correlated with wearing glasses---for example, because people often wear both hats and sunglasses outside and take them off inside. This correlation may be reflected in the training data, and a classifier trained to recognize a hat may rely on the presence of glasses. Consequently, the classifier may fail to recognize a hat in the absence of glasses, and vice versa.

\begin{figure}[t!]
\centering
\includegraphics[width=0.9\linewidth]{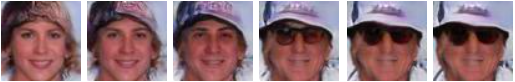} \\
\includegraphics[width=0.9\linewidth]{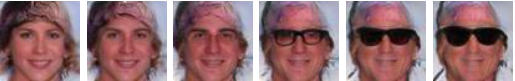}
\caption{Consider a GAN trained on a biased real-world dataset of faces where the presence of hats is correlated with the presence of glasses. Naively moving in a direction that adds glasses also adds a hat (\emph{Top}). We learn a direction in the latent space that allows us to add glasses, while not adding a hat (\emph{Bottom}). Note that attributes apart from the target attribute can change.}
\label{fig:changeGlasses}
\end{figure}

We propose using a GAN to generate more images with hats but not glasses and images with glasses but not hats, such that \texttt{WearingHat} is de-correlated from \texttt{Glasses} in the training data, by making perturbations in the latent space. 
Building on work by Denton et al.~\cite{DHMG19}, which demonstrates a method for learning interpretable image manipulation directions, we propose an improved latent vector perturbation method that allows us to preserve the \texttt{WearingHat} attribute while changing the \texttt{Glasses} attribute (Figure~\ref{fig:changeGlasses}).

\smallsec{Protected attributes} Our goal is to examine and mitigate biases of sensitive attributes such as gender expression, race, or age in visual classifiers. However, visual manipulations or explicit classifications along these dimensions have the potential to perpetuate harmful stereotypes (see ~\cite{google_VB}). Hence in our illustrations, we use \texttt{Glasses} as the protected attribute, as it has a clear visual signal. % that is less harmful to infer. 
In the quantitative experimental results, we report our findings on the more sensitive protected attributes of gender expression and age. 

\smallsec{Contributions}
We propose a method for perturbing vectors in the GAN latent space that successfully de-correlates target and protected attributes and allows for generating a de-biased dataset, which we use to augment the real-world dataset.
%augmenting and de-biasing the real-world dataset.
Attribute classifiers trained with the augmented dataset achieve quantitative improvements in several fairness metrics over both baselines and prior work~\cite{SHCV19FairnessGAN,Sharmanska2020contrastive,WQKGNHR19DomainBiasMitigation}, while maintaining comparable average precision. 
Furthermore, we analyze the CelebA~\cite{LLWT15CelebA} attributes with respect to label characteristics\footnote{We observe several discrepancies in the CelebA~\cite{LLWT15CelebA} attribute labels and categorize the attributes into three categories: inconsistently labeled, gender-dependent, and gender-independent.}, discriminability, and skew, and discuss how these factors influence our method's performance. 
We also evaluate our design choices with ablation studies and the results demonstrate the effectiveness of our augmentation method.\footnote{Code for all our experiments can be found at \url{https://github.com/princetonvisualai/gan-debiasing}.}

\section{Related Work}

\smallsec{De-biasing models}
The effect of gender and racial bias on AI models has been well documented~\cite{BCZSK16DebiasWord,BG18IntersectionalDataset,HBSDR18BiasCaptioning,WZYCO19BalancedDatasets,WQKGNHR19DomainBiasMitigation}. 
Models trained on biased data sometimes even amplify the existing biases~\cite{ZWYOC17BiasAmp}.
Tools such as AI Fairness 360~\cite{aif360-oct-2018} and REVISE~\cite{wang2020revise} surface such biases in large-scale datasets and enable preemptive analysis.
In parallel, various work propose methods for mitigating unwanted dataset biases from influencing the model.
Oversampling techniques~\cite{bickel09discriminative,elkan01CostSensitiveLearning} duplicate minority samples in imbalanced data to give them higher weight in training.
Some work propose to mitigate bias through adversarial learning~\cite{WZYCO19BalancedDatasets,ZLM18AdversarialLearning} or through learning separate classifiers for each protected attribute~\cite{RAM17inclusivefacenet,WQKGNHR19DomainBiasMitigation}.
Other work improve fairness by introducing constraints~\cite{lokh2020fairalm} or regularization terms~\cite{Baharlouei_ICLR2020} during training.
Contrary to these algorithmic approaches, our work aims to mitigate biases by training the model with a generated de-biased dataset.

\smallsec{Generating and perturbing images using GANs} 
Generative Adversarial Network (GAN)~\cite{GPMXWOCB14GANs} is a popular class of generative models composed of a generator and a discriminator trained in an adversarial setting.
Over the past few years, a number of works \cite{GAADC17WassersteinTraining,KALL17PGAN,KLA19StyleGAN,liu2020selfconditioned,SGZCRC16TrainingGAN} improved GANs to generate more realistic images with better stability. 
Shen et al.~\cite{SGTZ20LatentSpaceGANs} show that the latent space of GANs have semantic meaning and demonstrate facial attributes editing through latent space manipulation.
Denton et al.~\cite{DHMG19} propose a method to evaluate how sensitive a trained classifier is to such image manipulations, and find several attributes that affect a smiliing classifier trained on CelebA.
Balakrishnan et al.~\cite{Balakrishnan2020transect} use GANs to generate synthetic images that differ along specific attributes while preserving other attributes, and use them to measure algorithmic bias of face analysis algorithms. 
Unlike~\cite{Balakrishnan2020transect,DHMG19} who use the GAN-generated images to evaluate models, our work uses these generated images to train better attribute classification models.

\smallsec{Using GANs to augment datasets} 
Several works use GANs to augment datasets for low-shot~\cite{HG17LowShotRecognition} and long-tail~\cite{ZLQL17EmotionCycleGAN} recognition tasks, whereas our work focuses specifically on de-biasing classifiers affected by dataset bias. 
More related to our work are~\cite{GCSE19FairModeling,SHCV19FairnessGAN,Sharmanska2020contrastive} which leverage GANs to generate less biased data.
Choi et al.~\cite{GCSE19FairModeling}, given access to a small, unlabeled, and unbiased dataset, detect bias in a large and potentially biased dataset, and learn a generator that generates unbiased data at test time. 
Sattigeri et al.~\cite{SHCV19FairnessGAN} train a GAN with a modified loss function to achieve demographic parity or equality of odds in the generated dataset. 
Sharmanska et al.~\cite{Sharmanska2020contrastive} use an image-to-image translation GAN to generate more minority samples and create a balanced dataset.
While~\cite{GCSE19FairModeling,SHCV19FairnessGAN,Sharmanska2020contrastive} require training a new GAN for each bias they want to correct, our method uses a single GAN trained on a biased dataset to augment all attributes.

\section{Method} \label{sec:method}

\begin{figure}[t!]
\centering
  \includegraphics[width=0.9\linewidth]{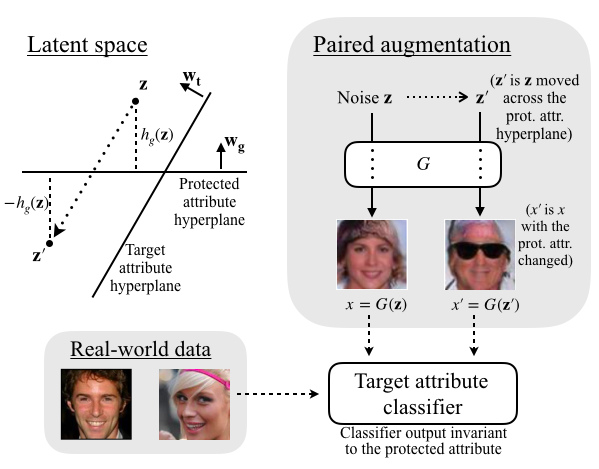}
%   \vspace*{-1.6pt}
\caption{\emph{(Top left)} Our latent vector perturbation method. 
%The trained GAN learns a distribution from which it samples $\z$. 
For each $\z$ sampled from the latent space of a trained GAN, we compute $\z'$ such that its target attribute score remains the same (according to $\wa$) while its protected attribute score is negated (according to $\wg$). \emph{(Top right)} We add images $G(\z)$ and $G(\z')$ to our training set, and train a target attribute classifier on both the real-world data and the generated de-biased data.}
\label{fig:Method}
\end{figure}

We study a class of problems where a protected attribute is correlated with a target label in the data $\mathcal{X}$, influencing target label prediction.
Let $t$ be the target label (e.g., \texttt{WearingHat} in the running example from Figure~\ref{fig:pullfig}) and $g$ be the protected attribute (e.g., gender expression or \texttt{Glasses} from our running example) with $t,g\in\{-1,1\}$. 
To mitigate the effect of unwanted dataset bias, we aim to generate a balanced set of synthetic images $\mathcal{X_\textit{syn}}$ where the protected attribute and target label are de-correlated.

Concretely, let $f_t$ be a function from images to binary labels that approximates the target label $t$, and $f_g$ be a function from images to binary labels that approximates the protected attribute $g$. 
We learn these classifiers in a supervised fashion with the original data.\footnote{$f_t$ is equivalent to the baseline classifier in Section~\ref{sec:baseline}.}
We now want to generate synthetic data $\mathcal{X_\textit{syn}}$ with the property that for $\mathbf{x} \in \mathcal{X_\textit{syn}}$:
\begingroup
\setlength\abovedisplayskip{3pt}
\setlength\belowdisplayskip{3pt}
\begin{equation}
    %P\left[f_t(x) = 1 | f_g(x) = 1 \right] = P\left[f_t(x) = 1 | f_g(x) = -1 \right] = P\left[f_t(x) =1   \right]
    P\left[{f_t}(\mathbf{x}) = 1 | {f_g}(\mathbf{x}) = 1 \right] = P\left[{f_t}(\mathbf{x}) =1   \right],
    \label{eq:indep}
\end{equation}
\endgroup
such that attributes $t$ and $g$ are de-correlated. 

\smallsec{De-biased dataset creation} 
To create $\mathcal{X_\textit{syn}}$, we use a GAN trained on real images $\mathcal{X}$ whose generator $G$ generates a synthetic image $\mathbf{x}$ from a random latent vector $\z \in \mathcal{Z}$. 
We can assign semantic attribute labels to these images using the learned functions ${f_t}(\mathbf{x})$ and ${f_g}(\mathbf{x})$. %and use them to augment the real-world dataset. 
However, as the GAN inherits correlations from its training data, a random sampling of $\z$ will produce an $\mathcal{X_\textit{syn}}$ with similar correlations and biases as $\mathcal{X}$. 
Hence, we propose a latent vector perturbation method that allows us to generate a de-biased $\mathcal{X_\textit{syn}}$.

We sample a random set of latent vectors $Z \subset \mathcal{Z}$ (inheriting the biases) and train classifiers $h_t, h_g \colon \mathcal{Z} \rightarrow [\num{-1}, 1]$  in the latent space that approximate ${f_t} \circ G$ and ${f_g} \circ G$, respectively.
That is, we train classifiers $h_t$ with input $\z$ and output ${f_t}(G(\z))$, and $h_g$ with input $\z$ and output ${f_g}(G(\z))$. 

Given a vector $\z$, we generate a complementary vector $\z'$ with the same (predicted) target label but the opposite (predicted) protected attribute label, or
\begingroup
\setlength\abovedisplayskip{3pt}
\setlength\belowdisplayskip{3pt}
\begin{equation}
\label{eq:pair_generation}
    h_t(\z') = h_t(\z), \; \; \; h_g(\z') = -h_g(\z).
\end{equation}
\endgroup
We note that this data generation method is agnostic to the type of classifier used to compute $h$. 

In our work, we assume that the latent spaces is approximately linearly separable in the semantic attributes, as observed and empirically validated by Denton et al.~\cite{DHMG19}. In this case, $h_t$ and $h_g$ can be represented as linear models (hyperplanes) $\wa$ and $\wg$ with intercepts $b_t$ and $b_g$ for the target and protected attributes respectively. We can derive a closed-form solution for $\z'$ as\footnote{Derivations are in the appendix (Section~\ref{sec:derivation}). 
$\|\wa\| = \|\wg\| = 1$.}
\begingroup
\setlength\abovedisplayskip{3pt}
\setlength\belowdisplayskip{3pt}
\begin{equation}
    \z' = \z - 2\left(\frac{\wgt \z +b_g}{1 - (\wgt\wa)^2}\right) \left(\wg - (\wgt\wa)\wa \right).
\end{equation}
\endgroup
%%%%%%%%%%%%%%%%%%%%%%%%%
This latent vector perturbation method is illustrated in Figure~\ref{fig:Method} (\emph{Top left}). A similar idea of hyperplane projection was presented in Zhang et al.~\cite{ZLM18AdversarialLearning}, although for a different goal of adversarial training. 
The sampling process results in a complementary image pair:
\begin{itemize}[topsep=1pt, itemsep=1pt, leftmargin=*]
\item $\mathbf{x}=G(\z)$ with target label ${f_t}(G(\z))$ and protected attribute label ${f_g}(G(\z))$
\item $\mathbf{x}'=G(\z')$ with target label ${f_t}(G(\z))$ and protected attribute label $-{f_g}(G(\z))$,
\end{itemize}
creating de-biased data $\mathcal{X}_{syn}$. We train our target attribute classifier with $\mathcal{X}$ and $\mathcal{X}_{syn}$, as shown in Figure~\ref{fig:Method}.

We label the generated images $\mathbf{x}$ and $\mathbf{x}'$ both with ${f_t}(\mathbf{x})$ because it allows us to capture the target attribute labels better than using ${f_t}(\mathbf{x})$ and ${f_t}(\mathbf{x}')$.
It is likely that the accuracy of ${f_t}$ is higher for the overrepresented group, and $\mathbf{x}$ will more often belong to the overrepresented group and $\mathbf{x}'$ to the underrepresented group.
However, other design choices are possible in our approach---for example, we could use $h_t(\z)$ and $h_t(\z')$ instead (after thresholding appropriately) or only use $\z$ for which ${f_t}(\mathbf{x}) = {f_t}(\mathbf{x}')$.
We compare these different design choices experimentally in Section~\ref{sec:designchoices}.

\smallsec{Advantages} 
Our data augmentation method has several attractive properties:
\begin{enumerate}[topsep=1pt, itemsep=1pt, leftmargin=*]
\item We use a single GAN trained on the biased real-world dataset to augment multiple target labels and protected attributes. This is in contrast to prior works like \cite{SHCV19FairnessGAN,GCSE19FairModeling} that require training a GAN for every pair of target and protected attributes. 
\item By augmenting samples $\z$ generated from (approximately) the original data distribution the GAN was trained on and maintaining their target attribute scores, our method preserves the intra-class variation of the images.
\item The samples $\z$ and $\z'$ are generated to simulate the independence goal of Equation~\ref{eq:indep}. By construction, $\z'$ maintains $\z$'s target label ${f_t}(G(\z))$ and takes on the opposite protected attribute label  $-{f_g}(G(\z))$.
\item Our method generalizes to multiple protected attributes $g$. We demonstrate how our method can simultaneously augment two protected attributes in Section~\ref{sec:comparisons_recent} when we compare our work to Sharmanska et al.~\cite{Sharmanska2020contrastive}. 
\end{enumerate}

\section{Experiments} \label{sec:exp}

In this section, we study the effectiveness of our data augmentation method on training fairer attribute classifiers. We first describe our experiment setup and compare our results to those of a baseline classifier. We then discuss how different factors influence our method's performance, and finally compare our work to several prior works.

\smallsec{Dataset and attributes categorization}
Given the task of training attribute classifiers that are not dependent on gender expression, we require a dataset that has target labels, as well as gender expression labels. 
CelebA~\cite{LLWT15CelebA} is a dataset with 2,022,599 images of celebrity faces, each with 40 binary attributes labels. 
We assume the \texttt{Male} attribute corresponds to gender expression.\footnote{Consistent with the dataset annotation and with the literature, we adopt the convention of using \texttt{Male} as our protected attribute. It is not clear if this label denotes assigned sex at birth, gender identity, or gender expression (socially perceived gender). Since the images were labeled by a professional labeling company~\cite{LLWT15CelebA}, we assume that the annotation refers to the perceived gender, or gender expression. Moreover, this attribute is annotated in a binary fashion. We would like to point out that none of these attributes (assigned sex at birth, gender identity, nor gender expression) are binary, however, we use these labels as is for our goal of de-biasing classifiers.} 
Among the other 39 attributes, we use 26 of them that have between 1\% and 99\% fraction of positive images for each gender expression.\footnote{We don't use \texttt{Blurry} as it has very few positive images ($\approx 5\%$).
We don't use \texttt{WearingNecklace} as the cropped images used in the GAN from \cite{PytorchPGAN} don't display the neck.} However, we noticed several discrepancies among the attribute labels, and decided to categorize the attributes into three categories: \textit{inconsistently labeled}, \textit{gender-dependent}, and \textit{gender-independent}. %In the results section, we discuss how our method's performance varies for these categories.

\begin{figure}
    \centering
    \includegraphics[width=0.9\linewidth]{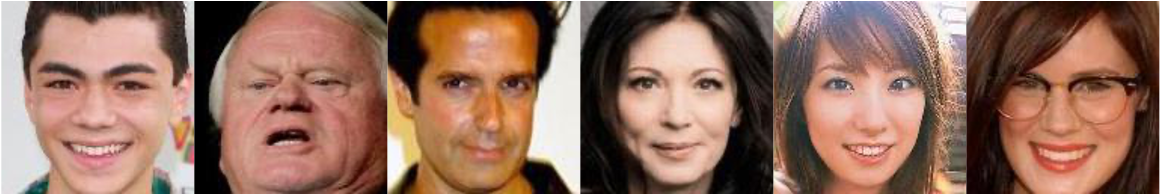} % straighthair2
    \caption{Examples of CelebA \texttt{StraightHair} labels. 
    Some of these are labeled as having \texttt{StraightHair} (1st, 3rd, 5th) and some as not (2nd, 4th, 6th).
    We deemed this attribute as \emph{inconsistently labeled}.}
    \label{fig:celeba_straighthair}
\end{figure}

We categorized attributes as \textit{inconsistently labeled} when we visually examined sets of examples and found that we often disagreed with the labeling and could not distinguish between positive and negative examples.
This category includes \texttt{StraightHair} shown in Figure \ref{fig:celeba_straighthair}, as well as \texttt{BigLips}, \texttt{BigNose}, \texttt{OvalFace}, \texttt{PaleSkin}, and \texttt{WavyHair}.\footnote{We note that for \texttt{BigNose}, we found that while there were some images that were easy to classify as having a big nose, or not having a big nose, most images were between these two extremes, and we believe that different annotators marked these `in-between' images differently. The same is true for the attribute \texttt{BigLips}.} While we report results on these attributes for completeness in Section~\ref{sec:baseline}, classifiers trained on these attributes may behave erratically. 

\begin{figure}
    \centering
    \includegraphics[width=0.9\linewidth]{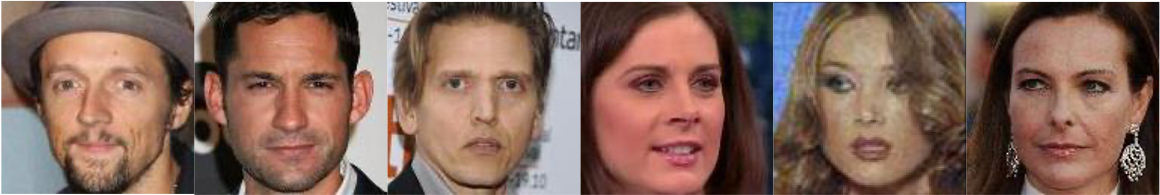} % young2
    \caption{Examples of CelebA \texttt{Young} labels.
    The first three images are labeled \texttt{Male}, \texttt{Young} while the last three images are labeled \texttt{not Male}, \texttt{not Young}, even though the first three appear older than the last three.
    We deemed this attribute as \emph{gender-dependent}.}
    \label{fig:celeba_young}
\end{figure}

Of the remaining attributes with more consistent labeling, we found that some attribute labels are \emph{gender-dependent}. That is, images are labeled to have (or not have) these attributes based on the perceived gender.
For example in Figure~\ref{fig:celeba_young}, we observe that the images labeled as \texttt{Young} and \texttt{Male} appear much older than the images labeled as \texttt{Young} and \texttt{not Male}. Other attributes in this category are \texttt{ArchedBrows}, \texttt{Attractive}, \texttt{BushyBrows}, \texttt{PointyNose} and \texttt{RecedingHair}. 

The \textit{gender-independent} attribute labels appear to be reasonably consistent among annotators, and do not appear to depend on the gender expression. We classified 14 attributes into this category: \texttt{Bangs}, \texttt{BlackHair}, \texttt{BlondHair}, \texttt{BrownHair}, \texttt{Chubby}, \texttt{Earrings}, \texttt{EyeBags}, \texttt{Glasses}, \texttt{GrayHair}, \texttt{HighCheeks}, \texttt{MouthOpen}, \texttt{NarrowEyes}, \texttt{Smiling}, and \texttt{WearingHat}. While we use the label `gender-independent' we note that these attributes can still be correlated with gender expression---for example \texttt{Earrings} are much more common among images labeled as \texttt{not Male} than those labeled as \texttt{Male}.

\smallsec{Implementation details}
To generate images, we use a Progressive GAN~\cite{KALL17PGAN} with a 512-D latent space trained on the CelebA~\cite{LLWT15CelebA} training set from the PyTorch GAN Zoo~\cite{PytorchPGAN}.
We use 10,000 synthetic images, labeled with baseline attribute classifiers, and learn hyperplanes ($h_t$, $h_g$) in the latent space with scikit-learn's~\cite{scikit-learn} linear SVM implementation.
%using the linear SVM implementation from the scikit-learn library~\cite{scikit-learn}.

For all attribute classifiers, we use ResNet-50~\cite{HZRS16ResNet} pre-trained on ImageNet~\cite{Russakovskyplus15Imagenet} as the base architecture. We replace the linear layer in ResNet with two linear layers with the hidden layer of size 2,048. Dropout and ReLU are applied between these. The inputs are $64{\times}64$ images and their target attribute labels. We train all models with the binary cross entropy loss for 20 epochs with a batch size of 32.
We use the Adam~\cite{KB14Adam} optimizer with a learning rate of 1e-4. We save the model with the smallest loss on a validation set that has the same distribution as the training set.

The baseline model is trained on the CelebA training set $\mathcal{X}$ with 162,770 images. Our model is trained on $\mathcal{X}$ and the balanced synthetic dataset $\mathcal{X}_{syn}$ (160,000 pairs of images).\footnote{We trained classifiers using different number of synthetic pairs for 4 different attributes, and found that AP stabilizes after 160,000 pairs, which is what we used to train our classifiers.} Results are reported on the CelebA test set unless noted otherwise. Error bars are 95\% confidence intervals estimated through bootstrapping. We note that we use a single GAN to construct the de-biased dataset for each target attribute, and then train separate classifiers for each target attribute. We also emphasize that protected attribute labels are only used in learning $h_g$ and in evaluation.%, never in training of the target attribute classifiers.

\smallsec{Evaluation Metrics}
We use \emph{average precision (AP)} to measure the accuracy of the classifiers. AP is a threshold-invariant accuracy metric that summarizes the precision and recall curve. We use this metric to ensure that our models learn a reasonable classification rule. AP, however, does not capture a classifier's behavior on different protected classes, and in fact, we expect to see a slight dip in overall AP when our model improves on some of the fairness metrics.

Multiple metrics have been proposed to measure fairness of a model~\cite{HPS16EqualityOdds,ZVGG17EqualityOpportunity,ZWYOC17BiasAmp,CKP09DemographicParity,chen2020riskdistribution} and each of these measures a different notion of fairness. 
In our work, we use three metrics for comprehensive understanding.
First, we measure the \emph{difference in equality of opportunity (DEO)}, i.e. the absolute difference between the false negative rates for both gender expression, as in Lokhande et al.~\cite{lokh2020fairalm}\footnote{In our experiments, we choose a calibrated threshold on the validation set, i.e, a threshold that ensures that we make the same number of positive predictions as the ground truth, to compute both DEO and BA. We tried other ways of choosing the threshold, such as choosing the one that gives the best $F_1$ score on a validation set, and while the values varied, they did not change our findings.}.

As our second fairness metric, we use the \emph{bias amplification (BA)} metric proposed by Wang and Russakovsky~\cite{wang2021directional}.
%\footnote{A recent submission shows that the metric proposed by Zhao et al.~\cite{ZWYOC17BiasAmp} does not take into account the base rate of target attribute for different protected groups or the direction of the bias amplification.} 
Intuitively, BA measures how much more often a target attribute is predicted with a protected attribute than the ground truth value. Let $P_{t|g}$ be the fraction of images with protected attribute $g$ that have target attribute $t$, ${P}_{\hat{t}| g}$ be the fraction of images with protected attribute $g$ that are predicted to have target attribute $t$, $P_{t,g}$ be the fraction of images with target $t$ and protected attribute $g$, and $P_{t}$ and $P_g$ be the fraction of images with attribute $t$ and $g$ respectively. For each pair of target and protected attribute values, we add $(P_{t|g} - P_{\hat{t}|g})$ if $P_{t,g}>P_{t}P_{g}$ and $-(P_{t|g} - P_{\hat{t}|g})$ otherwise. A negative value implies that bias now exists in a different direction than in the training data. 

Both DEO and BA fluctuate based on the chosen classification threshold. 
Hence, as our final fairness metric, we use a threshold-invariant metric that measures the \emph{divergence between score distributions (KL)}~\cite{chen2020riskdistribution} defined as follows: Suppose $s_{g,t}$ represents a smoothed histogram of classifier scores of a certain protected attribute label and a target label, 
appropriately normalized as a probability distribution of the scores. For each target attribute label $t$, %\in\{\num{-1},1\}$, 
we measure $KL\big[s_{g=\num{-1},t}\|s_{g=1,t}\big] + KL\big[s_{g=1,t}\|s_{g=\num{-1},t}\big]$. That is, we measure the divergence of $g{=}\num{-1}$ and $g{=}1$ score distributions, separately for positive and negative attribute samples. This is a stricter notion of \emph{equalized odds}\cite{HPS16EqualityOdds}.

\subsection{Comparison with the baseline} \label{sec:baseline}
To start, we compare our model (i.e. target classifiers trained using both the balanced synthetic datasets $\mathcal{X}_{syn}$ and the real dataset $\mathcal{X}$) with a baseline model trained using just $\mathcal{X}$.
In Table~\ref{tab:baseline}, we show results on the four metrics, averaged for each of the three attribute categories.
As expected, our model performs better on all three fairness metrics, DEO, BA and KL, while maintaining comparable AP. For gender-independent attributes, AP drops from 83.9 to 83.0, while DEO improves from 16.7 to 13.9, BA improves from 0.3 to 0.0 and KL improves from 1.1 to 0.9. 
For gender-dependent attributes, the fairness metrics improve over the baseline, but the improvements are smaller compared to those of gender-independent attributes. 
Later in Section~\ref{sec:extensions}, we demonstrate an extension of our augmentation method with an improved performance on the gender-dependent attributes.   

Additionally, we conduct score change evaluations suggested by Denton et al.~\cite{DHMG19} and measure the change in target attribute score as we perturb the protected attribute in images. Specifically, we measure the classifier score difference between $G(\z)$ and $G(\z')$. This evaluation helps understand how the protected attribute influences a trained classifier's output.
We find that the model trained with our augmentation method consistently has a smaller change in score than the baseline: 0.09 vs. 0.12 for inconsistently labeled, 0.07 vs. 0.11 for gender-dependent, and 0.06 vs. 0.09 for gender-independent attributes. 
We also observe that the baseline score changes are higher when we try to construct underrepresented samples. 
Consider the attribute \texttt{ArchedBrows} where only 2.3\% of the training set images are labeled to have \texttt{ArchedBrows}, and appear masculine.
When we construct a $\z'$ with this target and protected value, the baseline classifier's score changes by 0.41. 
On the other hand, when we try to construct an image that is without \texttt{ArchedBrows} and appears feminine, which comprises 33.7\% of the training set, the baseline classifier score only changes by 0.094. This could be due to the errors that the baseline classifier makes on underrepresented images during synthetic image labeling, or could imply that underrepresented attributes are harder to maintain during image manipulations. 

\begin{table}[t!]
\centering

\resizebox{\linewidth}{!}{% Please add the following required packages to your document preamble:

\begin{tabular}{|c|c c || c c|}
% \cline{2-4}
% & \multicolumn{1}{|c}{Baseline} & Ours &  {\sc model\_inv} \\ \hline 
% \multicolumn{4}{|l|}{AP $\uparrow$}\\ \hline
\hline
\multirow{2}{*}{Attr. type}& \multicolumn{2}{c||}{AP $\uparrow$} & \multicolumn{2}{c|}{DEO $\downarrow$} \\
\cline{2-5} 
 & Baseline & Ours & Baseline & Ours \\ \hline
\multicolumn{1}{|c|}{Incons.} & \textbf{66.3 $\pm$ 1.8} & 65.2 $\pm$ 1.9 & 21.5 $\pm$ 4.4 & \textbf{16.5 $\pm$ 4.2} \\
\multicolumn{1}{|c|}{G-dep} & \textbf{78.6 $\pm$ 1.4} & 77.8 $\pm$ 1.4 & 25.7 $\pm$ 3.5 & \textbf{23.4 $\pm$ 3.6} \\
\multicolumn{1}{|c|}{G-indep.} & \textbf{83.9 $\pm$ 1.5} & 83.0 $\pm$ 1.6 & 16.7 $\pm$ 5.0 & \textbf{13.9 $\pm$ 5.2} \\
\hline 
\multirow{2}{*}{Attr. type}& \multicolumn{2}{c||}{BA $\downarrow$} & \multicolumn{2}{c|}{KL $\downarrow$} \\
\cline{2-5} 
& Baseline & Ours & Baseline & Ours \\ \hline
\multicolumn{1}{|c|}{Incons.} & 2.1 $\pm$ 0.6 & \textbf{0.5 $\pm$ 0.6} & 1.7 $\pm$ 0.3 & \textbf{1.3 $\pm$ 0.4} \\
\multicolumn{1}{|c|}{G-dep} &  2.3 $\pm$ 0.5 & \textbf{1.6 $\pm$ 0.5} & 1.3 $\pm$ 0.2 & \textbf{1.2 $\pm$ 0.2}\\
\multicolumn{1}{|c|}{G-indep.} & 0.3 $\pm$ 0.6 & \textbf{0.0 $\pm$ 0.5} &  1.1 $\pm$ 0.5 & \textbf{0.9 $\pm$ 0.6}\\
\hline 
\end{tabular}

}
\caption{
Comparison of our model (i.e. attribute classifier trained with our data augmentation method) to the baseline model. Arrows indicate which direction is better. Numbers are averages over all attributes within the specific category. As expected, we have slightly lower AP than the baseline, but perform better on the three fairness metrics, DEO, BA, and KL.}
\label{tab:baseline}
\end{table}

We next examine several factors that could influence our method, including how easy the protected attribute is to learn compared to the target attribute and how data skew affects our method. We discuss the former here and provide more information about the latter in the appendix (Section~\ref{sec:factors}).

\smallsec{Discriminability of attributes}
Nam et al.~\cite{nam2020learning} recently observed that correlations among attributes affect a classifier only if the protected attribute is `easier' to learn than the target attribute. 
Inspired by their observation, we conduct a two-step experiment to understand how the relative discriminability of attributes affects our method's effectiveness.

First, we put a pair of CelebA attributes in competition to assess their relative discriminability. Experiment details are in the appendix. We find that gender expression is one of the easiest attributes to learn (\texttt{Gender} is easier than all but \texttt{Glasses} and \texttt{WearingHat}), which may be why gender bias is prevalent in many models. On the other hand, \texttt{Young} is relatively hard for a model to learn (\texttt{Young} is harder to learn than all but 4 other attributes), so its correlation with other attributes may not be as influential. 

\begin{table}[t!]
\centering
\resizebox{\linewidth}{!}{
\begin{tabular}{|c|cc|cc|cc|}
\hline
\multirow{3}{*}{\begin{tabular}[c]{@{}c@{}}Protected\\ Attribute\end{tabular}} & \multicolumn{6}{c|}{Improvement over baseline $\uparrow$} \\
\cline{2-7}
 & \multicolumn{2}{c|}{DEO}  & \multicolumn{2}{c|}{BA} & \multicolumn{2}{c|}{KL} \\
 \cline{2-7}
 & Easy & Hard  & Easy & Hard & Easy & Hard \\
 \hline
{\tt Glasses} (0,19) & -- & \textbf{4.1}  & -- & \textbf{0.9} & -- & \textbf{0.0} \\
{\tt Gender} (2, 17) & 0.8 & \textbf{3.2}  & 0.0 & \textbf{0.4} & -0.2 & \textbf{0.2} \\
{\tt Young} (15, 4) & -0.2 & \textbf{2.1}  & 0.2 & \textbf{1.0} & -0.2 & \textbf{0.0} \\ \hline
\end{tabular}}
\caption{Improvement over baseline for different fairness metrics when using different protected attributes. Next to the protected attribute are numbers of attributes that are `easier' and `harder' to learn, compared to the protected attribute. Columns `Easy' (`Hard') show the averages of all non-inconsistent target attributes that are easier (harder) for a classifier to learn. We note that our method works better when the target attribute is `harder' to learn.}
\label{tab:other_prot_attributes}
\end{table}

Next, to understand how the relative discriminability of attributes affects our method's performance, we train target attribute classifiers for gender-dependent and gender-independent attributes, using \texttt{Young} and \texttt{Glasses} as protected attributes.
In Table~\ref{tab:other_prot_attributes}, we report our method's improvement over baseline in the three fairness metrics. 
For each protected attribute, we report the average improvement separately for `easier' and `harder' target attributes. While training with our augmentation method generally outperforms the baseline on the three fairness metrics, as expected, the improvement is greater for target attributes that are harder to learn than the protected attribute, for example, for \texttt{Young}, the improvement in DEO over baseline is -0.2 for easy target attributes, and 2.1 for hard target attributes.

\smallsec{Skew of the dataset} The \emph{skew} of a target attribute $t$ is measured following the literature~\cite{WQKGNHR19DomainBiasMitigation} as $\frac{\max (P_{-1}, P_1)}{P_{-1}+P_1}$ where $P_{-1}$ is the number of images with $t{=}1$ and protected attribute label $g{=}-1$, and $P_1$ is the number of images with $t{=}1$ and protected attribute label $g{=}1$. We find that our augmentation method is most effective on attributes with low to moderate skew. %, and less on attributes that are severely skewed. 
Full details are in the appendix.

\subsection{Ablation studies}\label{sec:designchoices}
We now examine the design choices made in our method.% through several ablation studies.

\smallsec{Removal of $\z'$ samples}
First, we evaluate the effect of $G(\z')$ on the classifier. We train a classifier with just $G(\z)$ and the real dataset $\mathcal{X}$, and compare its performance against the performance of our model, trained with $G(\z)$, $G(\z')$, and $\mathcal{X}$ on the gender-dependent and gender-independent attributes.
While the new classifier's AP is higher than that of our model (82.9 vs. 82.6), all fairness metrics are worse: DEO is higher (19.7 vs. 16.1), BA is higher (1.1 vs. 0.5) and KL is higher (1.6 vs 1.3). All numbers were calculated on the validation set. In fact, it performs worse on the fairness metrics than the baseline model trained on $\mathcal{X}$. This result suggests that simply synthesizing more images with a GAN and adding them to the training data does not improve the model but rather hurts performance. Possible reasons include the image and label noise of $G(\z)$ and the skew of $G(\z)$ being worse than the original data the GAN was trained on. The fairness metrics improve only when we add $G(\z')$, and make the training data more balanced.

\begin{figure}[t!]
\centering
\resizebox{\linewidth}{!}{
\begin{tabular}{c p{7pt}|c|cc|}
\cline{3-5}
\multirow{6}{*}{\includegraphics[width=0.45\linewidth]{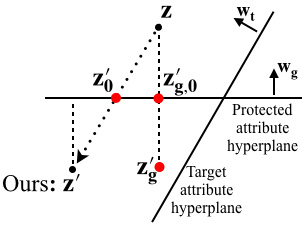}} && \multirow{2}{*}{Perturbation} & \multicolumn{2}{c|}{AP $\uparrow$} \\
\cline{4-5}
& & & G-dep & G-indep \\
\cline{3-5}
& & $\textbf{z}'_{g,0}$ & 74.0 & 79.9\\
& & $\textbf{z}'_{g}$ &  69.6 & 77.3\\
& & $\textbf{z}'_{0}$ & 74.4 & 79.8\\  
& & $\textbf{z}'$ (ours) & \textbf{76.0} & \textbf{81.4}\\
\cline{3-5}\\
\end{tabular}
}
\caption{Comparison of different perturbation choices. We train attribute classifiers using only synthetic images generated from the perturbations, and measure the mean AP over all target attributes on the validation set. The classifier trained with $\z'$ (our choice) has the highest AP.}
\label{fig:AvgPrecFakeOnly}
\end{figure}

\smallsec{Choice of $\z'$}
Next, we evaluate our choice of $\z'$ through examining a number of alternative perturbation choices visualized in Figure~\ref{fig:AvgPrecFakeOnly}. We train classifiers on just the generated data for gender-dependent and gender-independent attributes and compare the overall AP on the validation set.
As expected, training with $\z'$ (our choice) 
has the highest AP.

\smallsec{Filtering $\z$'s and using different labels for synthetic images}
Since we hallucinate labels for the synthetic images, some of these labels may be incorrect and harm our classifier. 
We try three different ways of addressing this issue: 
First, we try learning hyperplanes with different fractions of positive and negative samples. We find that while this improves the hyperplane accuracy, the downstream classifiers trained with samples generated using different hyperplanes have similar performances. 
%Full results are in the supplementary material. %(Section~\ref{sec:extra_ablations}). 
For the second and third methods, we use the original hyperplanes learned in our method, but vary the vectors/labelling used. We remove points that are incorrectly classified by the baseline classifier after perturbing the latent vector from $\z$ to $\z'$, i.e, we remove all points wherein $f_t(G(\z)) \not= f_t(G(\z'))$, and use the remaining synthetic images and the real dataset to train the classifiers.
Third, we label the synthetic images $G(\z)$ and $G(\z')$ with $h_t(\z)$, and use these labels to train the classifiers. 
We compare their performance to our method on the validation set.
We find that these two methods result in a slight drop in AP (79.8 when using $h_t$ scores, 82.1 when removing incorrectly classified points, and 82.6 for our method), as well as a small drop in the fairness metrics (the average DEO is 18.1 when using $h_t$ scores, 17.4  when removing incorrectly classified points, and 16.1 for our method), suggesting that our current labeling of the synthetic images works well. {Full results are in the appendix (Section~\ref{sec:extra_ablations})}. 
%We try different ways of addressing this issue: First, we remove 

\subsection{Comparison with prior work}
\label{sec:comparisons_recent}
In this section, we compare our method to few recent works~\cite{SHCV19FairnessGAN,Sharmanska2020contrastive,WQKGNHR19DomainBiasMitigation}.
One of the current challenges in the space of AI fairness is the lack of standardized benchmarks and metrics. 
While some of this stems from the complexity of the problem at hand (where it is difficult and even counter-productive to use a single fairness definition), in the computer vision community, we believe that more effort should be made to provide thorough comparison between methods. Each work we consider here uses slightly different evaluation protocols and benchmarks. We made comparisons to the best of our ability, and hope that our work helps enable more standardization and empirical comparisons.

\smallsec{Fairness GAN} Sattigeri et al.~\cite{SHCV19FairnessGAN} use GANs to create datasets that achieve either demographic parity (Dem. Par.) or equality of opportunity (Eq. Opp.). They train classifiers for the \texttt{Attractive} attribute on just the generated data, using gender expression as the protected attribute. We train classifiers with our pair-augmented synthetic data to mimic the conditions of Fairness GAN, and evaluate both on the CelebA test data. Comparison results are in Table~\ref{tab:FairnessGANcomp}. Our model performs better on most metrics, even though we use a single GAN to augment all attributes.

\begin{table}[t]
\centering
\resizebox{\linewidth}{!}{
\begin{tabular} {c|cc|cc|cc|}
  \cline{2-7}
      & \multicolumn{4}{c|}{Fairness GAN \cite{SHCV19FairnessGAN}}  & \multicolumn{2}{c|}{Ours} \\ \cline{2-5}
      & \multicolumn{2}{c|}{Dem. Par.} & \multicolumn{2}{c|}{Eq. Opp.} &\multicolumn{2}{c|}{(Synthetic only)} \\ 
\hline
\multicolumn{1}{|c|}{Gender exp. $g$} & $g{=}\num{-1}$ & $g{=}1$ & $g{=}\num{-1}$ & $g{=}1$ & $g{=}\num{-1}$ & $g{=}1$\\ \hline
\multicolumn{1}{|c|}{FPR $\downarrow$} & 0.52 & 0.26 & 0.42 & \textbf{0.17} & \textbf{0.22} & {0.39} \\
\multicolumn{1}{|c|}{FNR $\downarrow$} & 0.18 & 0.41 & 0.21 & 0.44 & \textbf{0.06} & \textbf{0.27} \\
\multicolumn{1}{|c|}{Error $\downarrow$} & 0.30  & 0.28  & 0.29 & 0.23 & \textbf{0.21} & \textbf{0.18} \\ \hline
\multicolumn{1}{|c|}{Error Rate $\downarrow$} & \multicolumn{2}{c|}{0.22} & \multicolumn{2}{c|}{0.29} & \multicolumn{2}{c|}{\textbf{0.20}}\\ \hline
\end{tabular}
}
\caption{Comparison of the \texttt{Attractive} classifier trained using synthetic data from Fairness GAN~\cite{SHCV19FairnessGAN} and the classifier trained using our pair-augmented synthetic data. The latter (ours) outperforms on most metrics.}
\label{tab:FairnessGANcomp}
\end{table}

\smallsec{Contrastive examples generated by image-to-image translation GANs}
Sharmanska et al.~\cite{Sharmanska2020contrastive} propose a different method for balancing a biased dataset using StarGAN~\cite{choi2018stargan}, a class of image-to-image translation GANs. They use two protected attributes, age and gender expression, and create a balanced dataset by creating contrastive examples, i.e. images of different ages and gender, for each image in the training set. They train a \texttt{Smiling} classifier with the augmented dataset, and propose making a prediction at test time only when the classifier makes the same prediction on the image and their contrastive examples. We extend our method to incorporate multiple protected attributes, and use gradient descent to find three points $\{z'_i\}_{i \in \{1, 2, 3\}}$ in the latent space that preserve the target attribute score and flip either the gender expression score, the age score, or both. This process gives us three synthetic images per training image, with which we train a \texttt{Smiling} classifier. To ensure that the error rates are similar across all four protected groups---(\texttt{Young}, \texttt{Male}), (\texttt{Young}, \texttt{not Male}), (\texttt{not Young}, \texttt{Male}), (\texttt{not Young}, \texttt{not Male})---they measure the the mean difference in the false positive and false negative rates between all pairs of protected groups. 
We reproduce their method to ensure that the results are reported on the same test set. We find that our model performs better in terms of the mean difference in FNR (0.34 versus their 0.54) and FPR (0.23 compared to their 0.46).  

\begin{table}[t!]
    \centering  
   \resizebox{\linewidth}{!}{
\begin{tabular}{|c|c|c|c|c|c|}
\hline
Skew & Method & AP $\uparrow$ & DEO $\downarrow$ & BA $\downarrow$ & KL $\downarrow$ \\
\hline
{Low/} & Dom. Ind. & \textbf{83.4 $\pm$ 1.3} & 7.0 $\pm$ 3.1 & \textbf{-0.1 $\pm$ 0.5} & 0.8 $\pm$ 0.7 \\
 Mod. & Ours & 81.4 $\pm$ 1.5 & \textbf{6.0 $\pm$ 3.0} & \textbf{-0.1 $\pm$ 0.5} & \textbf{0.3 $\pm$ 0.1}\\\hline
\multirow{2}{*}{High} & Dom. Ind. & \textbf{80.7 $\pm$ 1.6} & \textbf{14.9 $\pm$ 5.6} & \textbf{-0.4 $\pm$ 0.5} & \textbf{0.8 $\pm$ 1.0} \\
 & Ours & 80.4 $\pm$ 1.5 & {23.9 $\pm$ 5.5} & {0.9 $\pm$ 0.4} & 1.5 $\pm$ 0.6\\
\hline
\end{tabular}}
    \caption{Comparison of our method with domain independent training~\cite{WQKGNHR19DomainBiasMitigation}. Numbers reported are the mean over all gender-dependent and gender-independent attributes on the test set. We note that we perform better than domain-independent training for attributes with low to moderate skew.}% for DEO and KL.}
    \label{tab:dom_ind}
\end{table}

\begin{table}[t!]
    \centering  
   \resizebox{\linewidth}{!}{
\begin{tabular}{|c|c|c|c|c|}
\hline
 Method & AP $\uparrow$ & DEO $\downarrow$ & BA $\downarrow$ & KL $\downarrow$ \\
\hline
%\multicolumn{5}{|l|}{\textbf{Low to moderate skew}}\\ \hline
Weighted & {79.6 $\pm$ 1.6} & \textbf{5.7 $\pm$ 4.2} & \textbf{-2.8 $\pm$ 0.5} & \textbf{0.5 $\pm$ 0.4}\\
Adversarial & 81.3 $\pm$ 1.6 & 23.9 $\pm$ 4.4 & 1.5 $\pm$ 0.5 & 0.6 $\pm$ 0.5 \\
Ours & \textbf{81.5 $\pm$ 1.5} & {16.7 $\pm$ 4.7} & {0.5 $\pm$ 0.5} & {1.0 $\pm$ 0.5}\\
\hline
\end{tabular}}
    \caption{Comparison of our method with weighted and adversarial training from~\cite{WQKGNHR19DomainBiasMitigation}. Numbers reported are the mean over all gender-dependent and gender-independent attributes on the test set. We note that the weighted model overall performs better on the fairness metrics, however, the large negative BA suggests that the model now has bias in the opposite direction, to the extent that the AP drops. The adversarial model performs significantly worse than ours on DEO and BA, and marginally better on KL.}
    \label{tab:weighted}
\end{table}

\smallsec{Effective training strategies for bias mitigation} 
Wang et al.~\cite{WQKGNHR19DomainBiasMitigation} quantitatively compare different techniques for bias mitigation, including weighted training~\cite{bickel09discriminative,elkan01CostSensitiveLearning}, adversarial training with losses inspired by~\cite{AZN18FairnessBlindness,ZLM18AdversarialLearning}, and their proposed \emph{domain discriminative} and \emph{domain independent} training.
We compare our method to their best performing domain independent training method where they learn separate classifiers for each protected attribute class and combine them to leverage any shared information. We report results for all gender-dependent and gender-independent attributes in Table~\ref{tab:dom_ind}. We find that our method performs better for attributes with low to moderate skew ($<$0.7)---DEO is 6.0 compared to 7.0, KL is 0.3 compared to 0.8---whereas domain independent training performs better for attributes with high skew---DEO is 23.9 compared to 14.9, KL is 1.5 compared to 0.8. 
This result is consistent with our earlier observation that our method works well for low to moderately skewed datasets. 
Wang et al also use a simpler weighted training method that reweights samples such that the protected attribute classes have equal weight and an adversarial training method that uses a minimax objective to maximize the classifier's accuracy on the objective while minimizing an adversary's ability to predict the protected attribute from the learned features. For weighted and adversarial training methods, we report results in Table~\ref{tab:weighted}. We find that while the weighted model overall performs well on the fairness metrics, it has a strongly negative BA (-2.7 versus our 0.5) indicating that bias is now in the opposite direction, 
and a low AP (79.6 versus our 81.5) suggesting that it makes incorrect predictions to reduce bias. For adversarial training, our method does better overall, with lower DEO (16.7 versus 23.9) and lower BA (0.5 versus 1.5).

\section{Extensions of our method}
\label{sec:extensions}
In this final section, we study two natural extensions of our method: using domain-dependent hyperplanes in place of the current domain-independent hyperplanes, and directly augmenting a real image dataset with GAN-inversion.

\smallsec{Domain-dependent hyperplanes}
Our method implicitly assumes the learned hyperplane $\wa$ behaves equally well for all $\z$, irrespective of the value of ${f_g}(G(\z))$.
However, for gender-dependent attributes, the hyperplane learned using samples with ${f_g}(G(\z)) {=}1$ may be very different from that learned using samples with ${f_g}(G(\z)){=} \num{-1}$. 

For these attributes, we extend our method to learn per-domain target attribute hyperplanes: 
$\mathbf{w}_{t_1}, b_{t_1}$ for points with ${f_g}(G(\z)){=}1$ and $\mathbf{w}_{t_{\num{-1}}}, b_{t_{\num{-1}}}$ for points with ${f_g}(G(\z)){=}-1$.
\begin{figure}[t]
    \centering
    \includegraphics[width=0.95\linewidth]{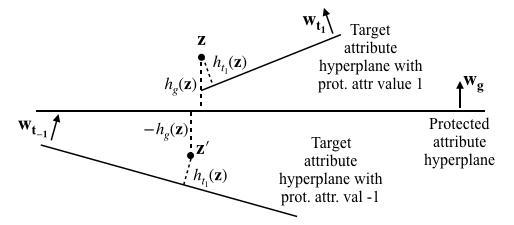}
    \caption{Computing $\z'$ %latent vectors for paired image augmentation 
    when the target attribute hyperplanes for each protected attribute class are very different.}
    \label{fig:per_gender}
\end{figure}
For $\z$ with $f_g(G(\z)){=}1$, we find $\z'$ such that 
\begingroup
\setlength\abovedisplayskip{3pt}
\setlength\belowdisplayskip{3pt}
\begin{equation}
\label{eq:per_gender}
\begin{split}
    \mathbf{w}_{t_{\num{-1}}}^T(\z')+b_{t_{\num{-1}}} & = \mathbf{w}_{t_1}^T(\z)+b_{t_1}, \mbox{ and}\\
\wgt\z'+b_g & = -\wgt(\z)-b_g 
\end{split}
\end{equation}
\endgroup
as shown in Figure \ref{fig:per_gender}. In order to compute $\z'$ that satisfies the above constraints, while minimizing $||\z - \z'||_2$, we note that all constraints are linear, hence the feasible region is the intersection of several hyperplanes. Starting from a point in this region, in each iteration, we find a new location of the point using gradient descent, then project it back onto the feasible region to maintain the constraints.

%We use stochastic gradient to find $\z'$ that satisfies the above constraints, while minimizing $||\z - \z'||_2$.

If $\mathbf{w}_{t_1}$ and $\mathbf{w}_{t_{\num{-1}}}$ are similar, these constraints are the same as Equation~\ref{eq:pair_generation} and this method of computing $\z'$ collapses to the first. 
We compare results of training a classifier that is augmented with images computed with domain-independent hyperplanes and with that using images computed with domain-dependent hyperplanes for all gender-dependent and gender-independent attributes over the validation set. We find that for gender-dependent attributes, using domain-dependent hyperplanes improves the fairness metrics considerably (DEO reduces from 21.4 to 17.2, BA reduces from 1.5 to 0.4, KL reduces from 1.2 to 1.0), without losing accuracy. However, for gender-independent attributes, we do not see significant improvement, suggesting that $\wa$ is similar to both $\mathbf{w}_{t_1}$ and $\mathbf{w}_{t_{\num{-1}}}$. Full results are in Table~\ref{tab:per_gender}. 
\begin{table}[t]
    \centering

\resizebox{\linewidth}{!}{
\begin{tabular}{|c|cc||cc|}
\hline
\multirow{2}{*}{Attr. type} & \multicolumn{2}{c||}{AP $\uparrow$} & \multicolumn{2}{c|}{DEO $\downarrow$}\\
\cline{2-5}
 & Dom-ind & Dom-dep & Dom-indep & Dom-dep \\ \hline
\multicolumn{1}{|c|}{G-dep} & \textbf{78.1 $\pm$ 1.5} & \textbf{78.1 $\pm$ 1.4} & {21.4 $\pm$ 4.0} & \textbf{17.2 $\pm$ 4.0}\\
\multicolumn{1}{|c|}{G-indep} & 84.5 $\pm$ 1.5 & \textbf{84.6 $\pm$ 1.6} & 13.9 $\pm$ 4.3 & \textbf{13.1 $\pm$ 4.6} \\ \hline
\multirow{2}{*}{Attr. type} & \multicolumn{2}{c||}{BA $\downarrow$} & \multicolumn{2}{c|}{KL $\downarrow$} \\
\cline{2-5}
& Dom-indep & Dom-dep & Dom-indep & Dom-dep \\ \hline
\multicolumn{1}{|c|}{G-dep} & 1.5 $\pm$ 0.5 & \textbf{0.4 $\pm$ 0.5}  & 1.2 $\pm$ 0.2 & \textbf{1.0 $\pm$ 0.3}  \\
\multicolumn{1}{|c|}{G-indep} & \textbf{0.1 $\pm$ 0.4}  & 0.2 $\pm$ 0.4 & \textbf{0.9 $\pm$ 0.5} & \textbf{0.9 $\pm$ 0.6}\\ \hline
\end{tabular}}
    \caption{Comparison of classifiers that use domain-dependent hyperplanes vs. domain-independent hyperplanes to compute $z'$. We see a significant improvement among Gender-dependent attributes when we use Domain-dependent hyperplanes. Numbers are reported on the validation set. }
    \label{tab:per_gender}
\end{table}

\smallsec{Augmenting real images with GAN-inversion} Our method operates in the GAN latent space and can only augment images that are generated from latent vectors, and so, only the GAN-generated images. Recently, several GAN-inversion methods have been proposed~\cite{Abdal_2019_ICCV,bau2019seeing,zhu2020indomain}. These methods invert a real image $\mathbf{x}_{real}\in\mathcal{X}$ to a vector $\z_{inv}$ in the latent space of a trained GAN. 
Using Zhu et al.~\cite{zhu2020indomain}, we tried directly augmenting the original dataset by perturbing $\z_{inv}$ to $\z'_{inv}$ with our method, creating $\mathbf{x}_{real}'{=}G(\z'_{inv})$ with the same target label and the opposite protected label of $\mathbf{x}_{real}$.
When we trained classifiers with datasets augmented in this way, however, we did not see an appreciable improvement, despite the more complex procedure (Table~\ref{tab:inverse}). 

\begin{table}[t]
    \centering
    \resizebox{\linewidth}{!}{
    \begin{tabular}{c|c|c|c|c|}
    \cline{2-5}
         & AP $\uparrow$ & DEO $\downarrow$ & BA $\downarrow$ & KL $\downarrow$ \\
    \hline
    \multicolumn{1}{|l|}{Without} & \textbf{82.6 $\pm$ 1.5} & 1.5 $\pm$ 2.3 & \textbf{1.3 $\pm$ 0.4} & \textbf{1.0 $\pm$ 0.5} \\
    \multicolumn{1}{|l|}{With inv.} & 82.4 $\pm$ 1.5 & \textbf{1.4 $\pm$ 2.3} & \textbf{1.3 $\pm$ 0.4} & \textbf{1.0 $\pm$ 0.5} \\ \hline
    \end{tabular}}
        \caption{Comparison of our classifiers (without) to classifiers trained using data augmented with a GAN-inversion module (with inv.). Numbers reported are the mean over all gender-dependent and gender-independent attributes on the validation set. We do not see an appreciable improvement.}
        %in using an inversion module.}
    \label{tab:inverse}
\end{table}

\section{Conclusions}

We introduced a GAN-based data augmentation method for training fairer attribute classifiers when correlations between the target label and the protected attribute (such as gender expression) might skew the results. We report results across a large number of attributes and metrics, including comparisons with existing techniques. We also analyze in detail when our method is the most effective. Our findings show the promise of augmenting data in the GAN latent space in a variety of settings. We hope our detailed analyses and publicly available code serve as a stepping stone for future explorations in this very important space.

\smallsec{Acknowledgements} This work is supported by the National Science Foundation under Grant No. 1763642 and the Princeton First Year Fellowship to SK. We also thank Arvind Narayanan, Deniz Oktay, Angelina Wang, Zeyu Wang, Felix Yu, Sharon Zhang, as well as the Bias in AI reading group for helpful comments and suggestions.

{\small
\bibliographystyle{ieee_fullname}
\bibliography{references}
}

\clearpage
\renewcommand{\thesubsection}{\Alph{subsection}}
%\section*{Fair Attribute Classification through Latent Space De-biasing (Appendix)}

\section*{Appendix}
%\olga{call it "appendix" (and refer to it as appendix throughout the paper). Number the sections differently from the main paper (e.g., appendix A, B, ... or A.1, A.2, ...)}

In this supplementary document, we provide additional details on certain sections of the main paper. 
\begin{itemize}[leftmargin=0pt, itemsep=1pt, topsep=1pt]
    \item[] \textbf{Section \ref{sec:derivation}:} We derive a closed form solution for $\z'$ which allows us to easily manipulate latent vectors in the latent space  (Section~\ref{sec:method}). 
    \item[] \textbf{Section~\ref{sec:dataset}} 
    We provide attribute-level results and further analysis of our main experiments (Section~\ref{sec:baseline}).
    \item[] \textbf{Section~\ref{sec:factors}:} 
    We discuss some factors that influence (or not) our method's effectiveness.
    \item[] \textbf{Section~\ref{sec:extra_ablations}:} 
    We provide more details on the ablation studies (Section~\ref{sec:designchoices}). 
    %\vikram{Let's not show images with manipulated gender}
    %\item[] \textbf{Section \ref{sec:gender_manip}:} 
    %We provide a few qualitative examples of gender expression manipulations that we did not include in the main paper due to their sensitive nature.
    \item[] \textbf{Section \ref{sec:choi}:} 
    We investigate how many images with protected attribute labels our method requires to achieve the desired performance.
\end{itemize}

\subsection{Derivation}
\label{sec:derivation}

%\vikram{Note: When we switched from using w_a to w_t, I just redefined \wa to be w_t}
In Section 3 of the main paper, we describe a method to compute perturbations within the latent vector space, such that the protected attribute score changes, while the target attribute score remains the same. More formally, if $h_t$ is a function that approximates the target attribute score, and $h_g$ is a function that approximates the protected attribute score, for every latent vector $\z$, we want to compute $\z'$ such that
\begin{equation}
\label{eq:target}
    h_t(\z') = h_t(\z), \; \; \; h_g(\z') = -h_g(\z).
\end{equation}
We assume that the latent space $\mathcal{Z}$ is approximately linearly separable in the semantic attributes. $h_t$ and $h_g$ thus can be represented as linear models $\wa$ and $\wg$, normalized as $||\wa|| = 1, ||\wg||=1$, for the target and protected attribute respectively, with intercepts $b_t$ and $b_g$. 

Equation \ref{eq:target} thus reduces to 
\begin{equation}
    \wat \z + b_t = \wat \z' + b_t, \; \; \; \wgt \z' + b_g = -\wgt \z - b_g.
\end{equation}
Simplifying, we get
\begin{equation}
    \wat (\z'-\z) = 0, \; \; \; \wgt (\z'+\z) + 2b_g = 0.
\end{equation}
These equations have infinitely many solutions, we choose the solution that minimizes the distance between $\z$ and $\z'$. This is true if $\z'-\z$ is in the span of $\{\wg, \wa\}$. Hence, we can represent $\z' - \z = \alpha \wa + \beta \wg$, and we get:
\begin{align}
     \wat (\z'-\z) & = 0 \\
     \wat (\alpha \wa + \beta \wg) & = 0\\
\Rightarrow \alpha =  - \beta \wat \wg\\
     \wgt ((\z'-\z) +2\z) + 2b_g & = 0\\
     \wgt (\alpha\wa + \beta \wg +2\z) + 2b_g & = 0\\
     -\beta (\wat \wg)^2 + \beta + 2\wgt\z + 2b_g & = 0\\
\Rightarrow (1-(\wat \wg)^2)\beta = -2(\wgt\z + b_g) \\
\Rightarrow \beta = -2\frac{(\wgt \z + b_g)}{(1-(\wat\wg)^2)}\\
\Rightarrow \alpha = 2\frac{(\wgt \z + b_g)(\wat\wg)}{(1-(\wat\wg)^2)}
\end{align}
This gives us a closed form solution for  $\z'$: 
\begin{equation}
    \z' = \z - 2\left(\frac{\wgt \z +b_g}{1 - (\wgt\wa)^2}\right) \left(\wg - (\wgt\wa)\wa \right).
\end{equation}

As a quick verification, we confirm that this value of $\z'$ maintains changes the protected attribute score, and maintains the target attribute score:

\begin{align*}
    &h_g(\z') \\
    &= \wgt \z' + b_g \\
    &= \wgt \left[\z  -2\left(\frac{\wgt \z +b_g}{1 - (\wgt\wa)^2}\right) \left(\wg - (\wgt\wa)\wa \right)\right]  + b_g \\
    &= \wgt \z - 2\left(\frac{\wgt \z +b_g}{1 - (\wgt\wa)^2}\right) \left(1  - (\wgt\wa)\wgt\wa \right) +b_g \\
    &= \wgt \z - 2(\wgt \z+b_g) +b_g  = -\wgt \z - b_g = -h_g(\z)
\end{align*}

\begin{align*}
	&h_a(\z') \\
	&= \wat \z' + b_t\\
    &= \wat\left[\z - 2\left(\frac{\wgt \z + b_g}{1 - (\wgt\wa)^2}\right) \left(\wg - (\wgt\wa)\wa \right) \right] +b_t\\
    &= \wat \z - 2\left(\frac{\wgt \z}{1 - (\wgt\wa)^2}\right) \left(\wat\wg - (\wgt\wa)\right) + b_t\\
    &= \wat \z + b_t = h_t(\z)
\end{align*}

\subsection{Attribute-level results}
\label{sec:dataset}

We provide attribute-level results and further analysis of our main experiments (Section 4.1 of the main paper).

% In this section we provide some analysis of the GAN latent space, investigating to what extent our assumption of linear separability of the latent space holds. We also evaluate how well we are able to maintain the target attribute, and other correlations between attributes that might affect our model. 

\subsubsection{Linear separability of latent space}
Our paired augmentation method assumes that the latent space is approximately linearly separable in the semantic attributes. Here we investigate to what extent this assumption holds for different attributes. 
As described in the main paper, the attribute hyperplanes were estimated with 10,000 samples using linear SVM.
% For each attribute, we estimate the target attribute hyperplane $h_a$ with 10,000 latent vectors and protected attribute class-specific hyperplanes $h_{a_1}$ and $h_{a_{-1}}$ with 15,000 latent vectors in total. 

In Table~\ref{tab:hyperplane-performance}, we report hyperplane accuracy and AP, measured on 160,000 synthetic samples, as well as the percentage of positive samples and the skew of the CelebA training set. The skew is calculated as $\frac{\max(N_{g=-1,a=1}, N_{g=1,a=1})}{N_{g=-1,a=1}+N_{g=1,a=1}}$ where $N_{g=-1,a=1}$ is the number of samples with protected attribute label $g{=-1}$ (perceived as not male) and target label 1 (positive) and $N_{g=1,a=1}$ defined likewise. The protected attribute class with more positive samples is noted in the skew column. 
We observe that most attributes are well separated with the estimated hyperplanes, except for those with high skew that have too few examples from underrepresented subgroups. 
% From this analysis, we can get an insight of how well our data augmentation method will work for a particular attribute.

For completeness, we also report our model's improvement over the baseline model on the four evaluation metrics. We did not find immediate correlations between the hyperplane quality with the downstream model performance.    

\begin{table*}[t]
\centering
\resizebox{\linewidth}{!}{
\begin{tabular}{|c|c|cc|cc|cc|rrrr|}
\hline
Attribute type & \multicolumn{3}{|c|}{Attribute statistics} & \multicolumn{2}{|c|}{Hyperplane acc.} & \multicolumn{2}{|c|}{Hyperplane AP} & \multicolumn{4}{|c|}{Improvement over baseline}\\ 
\hline
Inconsistently labeled & Positive & \multicolumn{2}{|c|}{Skew} & $g{=}{-1}$ & $g{=}{-1}$ & $g{=}{-1}$ & $g{=}1$ & AP & DEO & BA & KL\\ \hline
\texttt{BigLips} & 24.1\% & 0.73 & $g{=}{-1}$ & 80.3 & 92.0 & 49.7 & 28.9 & -0.35 & -0.79 & 1.23 & -0.03 \\
\texttt{BigNose} & 23.6\% & 0.75 & $g{=}1$ & 91.7 & 74.5 & 51.1 & 82.4 & -0.66 & 11.03 & 2.52 & 1.04 \\
\texttt{OvalFace} & 28.3\% & 0.68 & $g{=}{-1}$ & 75.4 & 74.2 & 85.3 & 63.1 & -1.82 & 7.53 & 3.33 & 0.77 \\
\texttt{PaleSkin} & 4.3\% & 0.76 & $g{=}{-1}$ & 94.4 & 96.9 & 48.4 & 30.9 & -1.90 & 4.26 & 0.31 & 0.26 \\
\texttt{StraightHair} & 20.9\% & 0.52 & $g{=}{-1}$ & 87.7 & 69.8 & 25.0 & 58.8 & -1.76 & 0.94 & 0.53 & -0.08 \\
\texttt{WavyHair} & 31.9\% & 0.81 & $g{=}{-1}$ & 73.0 & 92.1 & 79.4 & 23.5 & -0.65 & 7.59 & 1.33 & 0.26 \\ \hline
Gender-dependent & Positive & \multicolumn{2}{|c|}{Skew} & $g{=}{-1}$ & $g{=}1$ & $g{=}{-1}$ & $g{=}1$ & AP & DEO & BA & KL\\ \hline
\texttt{ArchedBrows} & 26.6\% & 0.92 & $g{=}{-1}$ & 72.3 & 92.1 & 82.6 & 25.5 & -0.69 & -3.31 & -0.09 & 0.02 \\
\texttt{Attractive} & 51.4\% & 0.77 & $g{=}{-1}$ & 88.4 & 81.0 & 97.9 & 81.9 & -0.33 & 3.25 & 0.98 & 0.41 \\
\texttt{BushyBrows} & 14.4\% & 0.71 & $g{=}1$ & 94.5 & 79.6 & 37.6 & 62.0 & -1.20 & 8.49 & 1.14 & 0.25 \\
\texttt{PointyNose} & 27.6\% & 0.75 & $g{=}{-1}$ & 73.6 & 82.9 & 84.4 & 59.9 & -1.32 & 3.25 & 0.99 & -0.40 \\
\texttt{RecedingHair} & 8.0\% & 0.62 & $g{=}1$ & 94.5 & 88.3 & 41.8 & 57.7 & -1.44 & 2.32 & 0.40 & 0.17 \\
\texttt{Young} & 77.9\% & 0.66 & $g{=}{-1}$ & 96.2 & 84.1 & 99.7 & 95.3 & -0.24 & 0.78 & 0.49 & 0.31 \\ \hline
Gender-independent & Positive & \multicolumn{2}{|c|}{Skew} & $g{=}{-1}$ & $g{=}1$ & $g{=}{-1}$ & $g{=}1$ & AP & DEO & BA & KL\\ \hline
\texttt{Bangs} & 15.2\% & 0.77 & $g{=}{-1}$ & 90.3 & 94.9 & 81.5 & 58.9 & -0.50 & 0.62 & 0.38 & 0.09 \\
\texttt{BlackHair} & 23.9\% & 0.52 & $g{=}1$ & 89.3 & 83.2 & 78.9 & 79.2 & -1.00 & 2.25 & 0.44 & 0.00 \\
\texttt{BlondHair} & 14.9\% & 0.94 & $g{=}{-1}$ & 88.9 & 97.1 & 82.7 & 19.8 & -0.77 & 1.04 & 0.23 & -0.12 \\
\texttt{BrownHair} & 20.3\% & 0.69 & $g{=}{-1}$ & 66.4 & 80.4 & 45.5 & 38.8 & -0.51 & -0.57 & -0.01 & 0.01 \\
\texttt{Chubby} & 5.8\% & 0.88 & $g{=}1$ & 99.1 & 89.9 & 7.6 & 33.8 & -1.95 & 4.08 & 0.01 & 0.13 \\
\texttt{EyeBags} & 20.4\% & 0.71 & $g{=}1$ & 90.7 & 74.4 & 64.1 & 74.4 & -1.74 & 8.30 & 1.91 & 0.58 \\
\texttt{Glasses} & 6.5\% & 0.80 & $g{=}1$ & 97.8 & 92.5 & 60.3 & 77.8 & -0.24 & -0.07 & 0.05 & -0.27 \\
\texttt{GrayHair} & 4.2\% & 0.86 & $g{=}1$ & 98.4 & 92.6 & 10.4 & 32.9 & -2.60 & 7.02 & 0.32 & 0.54 \\
\texttt{HighCheeks} & 45.2\% & 0.72 & $g{=}{-1}$ & 86.3 & 86.3 & 95.2 & 83.5 & -0.33 & -1.06 & 0.24 & 0.04 \\
\texttt{MouthOpen} & 48.2\% & 0.63 & $g{=}{-1}$ & 88.6 & 87.0 & 96.4 & 93.1 & -0.08 & 0.69 & 0.34 & -0.03 \\
\texttt{NarrowEyes} & 11.6\% & 0.56 & $g{=}{-1}$ & 93.8 & 92.1 & 29.6 & 26.4 & -0.97 & 3.10 & -0.53 & 0.12 \\
\texttt{Smiling} & 48.0\% & 0.65 & $g{=}{-1}$ & 91.5 & 90.7 & 98.0 & 96.5 & -0.09 & 1.01 & 0.67 & 0.03 \\
\texttt{Earrings} & 18.7\% & 0.97 & $g{=}{-1}$ & 71.8 & 96.3 & 56.9 & 3.0 & -0.63 & 8.18 & 0.64 & 1.40 \\
\texttt{WearingHat} & 4.9\% & 0.70 & $g{=}1$ & 97.4 & 94.0 & 45.0 & 60.6 & -0.95 & 2.67 & 0.14 & -0.06 \\ \hline
\textbf{Average} & 24.1\% & 0.73 &  & 87.4 & 86.9 & 62.9 & 55.7 & -0.95 & 3.18 & 0.69 & 0.21 \\ \hline
\end{tabular}
}
\caption{Attribute-level information. The columns are (from left to right) target attribute name, percentage of positive samples, skew, hyperplane accuracy, hyperplane AP, and our model's improvement over the baseline model on the four evaluation metrics.} 
\label{tab:hyperplane-performance}
\end{table*}

\subsubsection{Changes in baseline score}
\begin{figure}[t!]
    \centering
    \includegraphics[width=\linewidth]{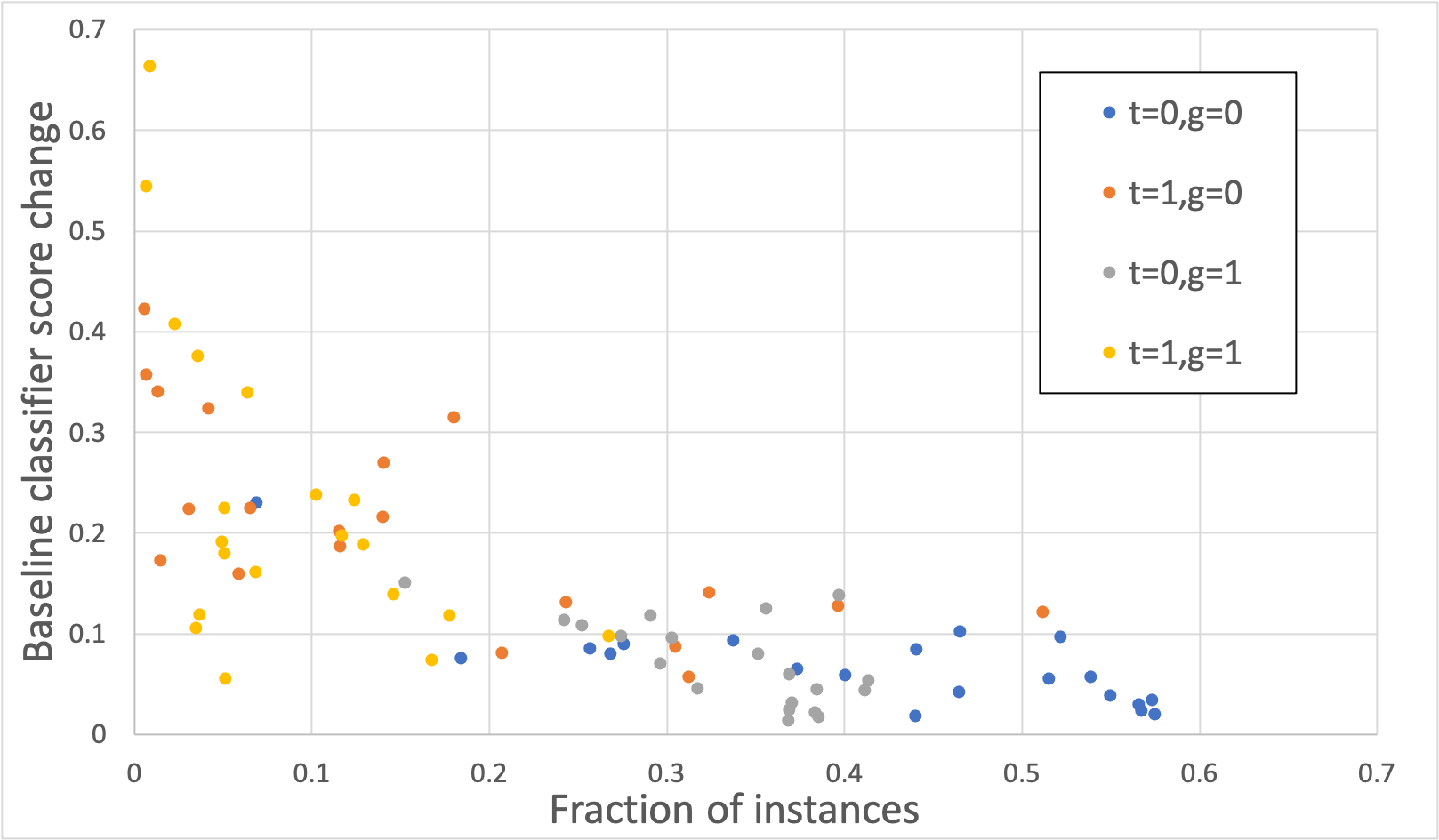}
    \caption{We plot average absolute change in the baseline classifier score versus the fraction of images in the dataset that have the corresponding ground truth labels. We separate them based on what the new ground truth values should be, for each attribute. We find that the score change is larger when creating an image with minority labels. This could be because we are unable to maintain the target attribute in this case or because the baseline classifier performs worse on minority images.}
    \label{fig:score_change}
\end{figure}

We next evaluate how well we are able to maintain the target attribute score when perturbing the latent vector. We use the change in the baseline classifier as a proxy to measure the target attribute score. We note that this measurement is flawed because the baseline classifier is known to perform worse on minority examples, however, we believe that this measurement still leads to some valuable insights. 
For each attribute, we measure the the absolute change in baseline score $|f_t(G(\z) - f_t(G(\z'))|$ over 5000 images, and compute averages based on what we expect the target and protected attribute values of $G(\z')$ to be. We plot this versus the fraction of images in the real world dataset that have these target and protected values (Figure~\ref{fig:score_change}). We find that there is a strong negative correlation. This could be because the target attribute is harder to maintain in this case, or because the baseline classifier has a tendency to misclassify minority samples. 

Another question that we were interested in was interactions between different attributes as we create balanced synthetic datasets for different attributes. We measured the change in baseline classifier score for different targets $t'$ when trying to maintain target attribute $t$ and found that some attributes changed drastically when creating a balanced dataset for any attribute (Table~\ref{tab:score_change_all}). For example, the attribute \texttt{Attractive} changed by a large amount irrespective of which target attribute we were trying to preserve. This suggests that some of these attributes are more sensitive to latent space manipulations.  

\begin{table}[t!]
\centering
\resizebox{\linewidth}{!}{\begin{tabular}{|c|c|c|c|}
\hline
Attribute & Change & Attribute & Change \\ \hline
\texttt{ArchedBrows} & \textbf{0.314} & \texttt{Glasses} & 0.109 \\
\texttt{Attractive} & \textbf{0.336} & \texttt{GrayHair} & 0.056 \\
\texttt{Bangs} & 0.120 & \texttt{HighCheeks} & \textbf{0.233} \\
\texttt{BlackHair} & 0.153 & \texttt{MouthOpen} & 0.187 \\
\texttt{BlondHair} & 0.180 & \texttt{NarrowEyes} & 0.066 \\
\texttt{BrownHair} & 0.158 & \texttt{PointyNose} & 0.152 \\
\texttt{BushyBrows} & 0.136 & \texttt{RecedingHair} & 0.069 \\
\texttt{Chubby} & 0.067 & \texttt{Smiling} & 0.176 \\
\texttt{Earrings} & 0.176 & \texttt{WearingHat} & 0.065 \\
\texttt{Eyebags} & \textbf{0.212} & \texttt{Young} & \textbf{0.268} \\ \hline
\end{tabular}}
\caption{We report the average classifier score change in an attribute when trying to create balanced datasets for other attributes. Classifier scores are between 0 and 1, and changes above 0.2 are bolded. We find that some attributes (e.g. \texttt{Attractive}, \texttt{Young}) change by a lot, whereas others (e.g. \texttt{GrayHair}, \texttt{WearingHat}) do not change much.}
\label{tab:score_change_all}
\end{table}

\subsection{Factors of influence}
\label{sec:factors}
% In this section, we provide more details for certain factors that influence our models performance (Section 4.1 of the main paper). 
In this section, we discuss in more detail how some factors influence (or not) our method's effectiveness (Section 4.1 of the main paper).  

\subsubsection{Skew of attributes}
For some attributes, the majority of the positive samples come from one gender expression. 
For example, \texttt{ArchedBrows} has a skew of 0.92 towards $g{=}\num{-1}$, that is, 92\% of positive \texttt{ArchedBrows} samples have gender expression label $g{=}\num{-1}$.
To understand the effect of data skew on our method's performance, we ran experiments with differently skewed data. From the 162,770 CelebA training set images, we created slightly smaller training sets where the attribute of interest (e.g. \texttt{HighCheeks}) has different values of skew. Specifically, we created three versions of training data each with skew 0.5, 0.7, 0.9, while keeping the total number of images fixed. We trained a GAN on each training set, created a synthetic de-biased dataset with our method, and trained an attribute classifier with the training set and 160,000 pairs of synthetic images. For comparison, we also trained baseline models on just the differently skewed training sets. The classifiers were evaluated on the CelebA validation set. Table~\ref{tab:skewdata} summarizes the results. Compared to the baseline, our model has lower AP as expected, better DEO for skew 0.5 and 0.7, worse DNAP, and better or on par BA. Overall, classifiers trained on more imbalanced data with higher skew perform worse on all metrics.

\begin{table}[ht!]
\centering

\resizebox{\linewidth}{!}{
\begin{tabular}{|c|cc|cc|}
\hline
\multirow{2}{*}{Skew} & \multicolumn{2}{c|}{AP $\uparrow$} & \multicolumn{2}{c|}{DEO $\downarrow$} \\ \cline{2-5} 
 & Base & Ours & Base & Ours   \\ \hline
\multicolumn{1}{|c|}{0.5} & \textbf{95.1 $\pm$ 0.3} & 93.6 $\pm$ 0.4 & 7.0 $\pm$ 1.7 & \textbf{6.6 $\pm$ 1.8} \\
\multicolumn{1}{|c|}{0.7} & \textbf{94.8 $\pm$ 0.3} & 94.1 $\pm$ 0.3 & 19.6 $\pm$ 1.9 & \textbf{19.4 $\pm$ 1.9} \\
\multicolumn{1}{|c|}{0.9} & \textbf{94.1 $\pm$ 1.7} & 93.1 $\pm$ 0.4 & \textbf{31.3 $\pm$ 2.0} & 32.9 $\pm$ 1.9  \\ \hline
\multirow{2}{*}{Skew}& \multicolumn{2}{c|}{BA $\downarrow$} & \multicolumn{2}{c|}{KL $\downarrow$} \\
 \cline{2-5} 
 & Base & Ours & Base & Ours  \\ \hline
\multicolumn{1}{|c|}{0.5} & -1.9 $\pm$ 0.5 & \textbf{-3.0 $\pm$ 0.5} & 0.4 $\pm$ 0.1 & \textbf{0.3 $\pm$ 0.1} \\
\multicolumn{1}{|c|}{0.7} & \textbf{3.4 $\pm$ 0.5} & \textbf{3.4 $\pm$ 0.5} & \textbf{0.9 $\pm$ 0.1} & \textbf{0.9 $\pm$ 0.1} \\
\multicolumn{1}{|c|}{0.9} & 7.1 $\pm$ 0.5 & \textbf{7.0 $\pm$ 0.5} & \textbf{1.7 $\pm$ 0.1} & 1.9 $\pm$ 0.1 \\ \hline
\end{tabular}
}
\caption{Comparison of \texttt{HighCheeks} attribute classifiers trained on differently skewed data.}
\label{tab:skewdata}
\end{table}

\subsubsection{Discriminability of attributes}

Nam et al.~\cite{nam2020learning} recently observed that correlations among attributes affect a classifier only if the protected attribute is `easier' to learn than the target attribute. 
Inspired by their observation, we design an experiment where we put a pair of CelebA attributes in competition to assess their relative discriminability. 
We create a fully skewed dataset in which half of the images have both attributes and the other half have neither. With this dataset, we train a classifier to predict if an image has both attributes or neither. At test time, we evaluate the classifier on a perfectly balanced subset of the CelebA validation set (where each of the four possible hat-glasses combinations occupies a quarter of the dataset), and compute AP for each attribute. If one attribute has a higher AP than the other, it suggests that this attribute is `easier' to learn than the other. We repeat this experiment with a second dataset skewed in a different way (i.e. half of the images have one attribute but not the other).

\begin{table}[t!]
    \centering
    \resizebox{\linewidth}{!}{
        \begin{tabular}{c | c c c c c c c c c c c c c c c c c c c c c c c c c c}

    & \rot{\texttt{ArchedBrows}} 
    & \rot{\texttt{Attractive}}
    & \rot{\texttt{Bangs}}
    %& \rot{\texttt{BigLips}}
    %& \rot{\texttt{BigNose}}
    & \rot{\texttt{BlackHair}}
    & \rot{\texttt{BlondHair}}
    & \rot{\texttt{BrownHair}}
    & \rot{\texttt{BushyBrows}}
    & \rot{\texttt{Chubby}}
    & \rot{\texttt{Earrings}}
    & \rot{\texttt{EyeBags}}
    & \rot{\texttt{Glasses}}
    & \rot{\texttt{GrayHair}}
    & \rot{\texttt{HighCheeks}}
    & \rot{\texttt{MouthOpen}}
    & \rot{\texttt{NarrowEyes}}
    %& \rot{\texttt{OvalFace}}
    %& \rot{\texttt{PaleSkin}}
    & \rot{\texttt{PointyNose}}
    & \rot{\texttt{RecedingHair}}
    & \rot{\texttt{Smiling}}
    %& \rot{\texttt{StraightHair}}
    %& \rot{\texttt{WavyHair}}
    & \rot{\texttt{WearingHat}}
    & \rot{\texttt{Young}} \\
    \hline
    \texttt{Gender} & \texttt{y} & \texttt{y} & \texttt{y} & % \texttt{y} & \texttt{y} & 
    \texttt{y} & \texttt{y} & \texttt{y} & \texttt{y} & \texttt{y} & \texttt{y} & \texttt{y} & \texttt{n} & \texttt{y} & \texttt{y} & \texttt{y} & \texttt{y} & %\texttt{y} & \texttt{y} &
    \texttt{y} & \texttt{y} & \texttt{y} & % \texttt{y} & \texttt{y} & 
   \texttt{n} & \texttt{y} \\
    \texttt{Glasses} & \texttt{y} & \texttt{y} & \texttt{y} & % \texttt{y} & \texttt{y} & 
    \texttt{y} & \texttt{y} & \texttt{y} & \texttt{y} & \texttt{y} & \texttt{y} & \texttt{y} & -- & \texttt{y} & \texttt{y} & \texttt{y} & \texttt{y} & %\texttt{y} & \texttt{y} & 
    \texttt{y} & \texttt{y} & \texttt{y} & %\texttt{y} & \texttt{y} & 
    \texttt{y} & \texttt{y} \\
    \texttt{Young} & \texttt{n} & \texttt{n} &  \texttt{n} & %\texttt{y} & \texttt{y} & 
    \texttt{n} & \texttt{n} & \texttt{y} & \texttt{n} & \texttt{n} & \texttt{y} & \texttt{n} & \texttt{n} & \texttt{n} & \texttt{n} & \texttt{n} & \texttt{n} & %\texttt{y} & \texttt{n} &
    \texttt{y} & \texttt{n} & \texttt{n} & %\texttt{y} & \texttt{y} & 
     \texttt{y} & -- \\
    \hline
    \end{tabular}
}
        \caption{Discriminability of attributes. We compare attributes on the row to those in the columns. \texttt{y} indicates that the attribute in the row is easier to learn than that in the column and \texttt{n} indicates the opposite. We find that gender expression is one of the easiest attributes to learn, while \texttt{Young} is relatively hard.}
    \label{tab:full_skew}
\end{table}

The results for gender-dependent and gender-independent attributes are in Table~\ref{tab:full_skew}. We report that an attribute is `easier' to learn than the other if it has a higher AP for both created datasets. We find that gender expression is one of the easiest attributes to learn, which may be why gender bias is prevalent in many models. On the other hand, \texttt{Young} is relatively hard for a model to learn, so its correlation with other attributes may not be as influential. 
We find that gender expression is one of the easiest attributes to learn (with gender expression having a higher AP than every attribute we tested except \texttt{WearingHat} and \texttt{Glasses}), which may be why gender bias is prevalent in many models. On the other hand, \texttt{Young} is relatively hard for a model to learn (\texttt{Young} is harder to learn than all but 4 other attributes), so its correlation with other attributes may not be as influential.

\subsection{Ablation studies}
\label{sec:extra_ablations}

In this section, we describe in more detail the ablation studies we have conducted to investigate how improved hyperplanes and use of different labels for synthetic images impact (or not) our method's performance (Section 4.2 of the main paper).

We first investigate if hyperplanes estimated with better balanced samples improve the performance of downstream attribute classifiers. 
We test this hypothesis by training models using hyperplanes that are estimated with different fractions of positive or negative samples. 

For the attribute \texttt{HighCheeks}, we estimate hyperplanes with different fractions of positive and negative samples, while keeping the total number of samples constant at 12,000 and the number of positive samples same for each gender expression.
We then train attribute classifiers with the CelebA training set and synthetic pair images augmented with these different hyperplanes. In Table~\ref{tab:underrep}, we report results evaluated on the CelebA validation set. 
We find that although the fairness metrics deteriorate as the target attribute hyperplanes were estimated with less balanced samples, this rate is relatively slow, and the downstream classifier still performs reasonably well.

\begin{table}[ht]
    \centering

    \resizebox{\linewidth}{!}{
    \begin{tabular}{|c|c|c|c|c|c|}
\hline
Fraction & AP $\uparrow$ & DEO $\downarrow$ & BA $\downarrow$ & KL $\downarrow$ \\ \hline
50.0\% & 95.1 $\pm$ 0.3 & 13.2 $\pm$ 1.7 & {0.5 $\pm$ 0.5} & 0.7 $\pm$ 0.1 \\
12.5\% & {95.1 $\pm$ 0.3} & 14.0 $\pm$ 1.7 & {0.8 $\pm$ 0.5} & \textbf{0.6 $\pm$ 0.1}\\
6.3\% & {95.1 $\pm$ 0.3} & {15.1 $\pm$ 1.8} & 1.3 $\pm$ 0.5 & 0.8 $\pm$ 0.2 \\
3.1\% & {95.1 $\pm$ 0.3} & {14.2 $\pm$ 1.7} & 1.0 $\pm$ 0.5 & 0.7 $\pm$ 0.1\\
1.6\% & 95.1 $\pm$ 0.3 & \textbf{12.9 $\pm$ 1.8} & \textbf{0.3 $\pm$ 0.5} & 0.7 $\pm$ 0.1 \\ \hline
\end{tabular}
}
    \caption{The amount of underrepresentation in samples used for hyperplane estimation doesn't appear to affect the performancee of the downstream classsification model much.}
    \label{tab:underrep}
\end{table}

Next, we tried training models with synthetic images with the same hallucinated target labels, i.e. using only $G(\z)$ and $G(\z')$ such that $f_t(G(\z)){=}f_t(G(\z'))$, and labeling synthetic images with $h_t(\z)$ in place of $f_t(G(\z))$.
Table~\ref{tab:h_scores_incorrect} contains all results. We report average results over all gender-dependent and gender-independent attributes. We find that both these ablations are comparable to ours, with in a slight loss in AP (79.8 and 82.1 versus 82.6), and worse fairness metrics in general (average DEO is 18.1 and 17.4  vs 16.1, BA is 0.9 and 0.7 vs 0.5). 

\begin{table}[t!]
\resizebox{\linewidth}{!}{
\begin{tabular}{c|c|c|c|c|}
\cline{2-5}
 & AP $\uparrow$ & DEO $\downarrow$ & BA $\downarrow$ & KL $\downarrow$ \\ \hline
\multicolumn{1}{|c|}{\begin{tabular}[c]{@{}c@{}}$f_t(G(\z)) =$\\ $\;\; f_t(G(\z'))$\end{tabular}} & 79.8 $\pm$ 1.6 & 17.4 $\pm$ 4.5 & 0.9 $\pm$ 0.4 & \textbf{1.0 $\pm$ 0.3} \\ \hline
\multicolumn{1}{|c|}{\begin{tabular}[c]{@{}c@{}}Labels \\ computed \\ using $h_t$\end{tabular}} & 82.1 $\pm$ 1.5 & 18.1 $\pm$ 4.2 & 0.7 $\pm$ 0.4 & 1.4 $\pm$ 0.8 \\ \hline
\multicolumn{1}{|c|}{Ours} & \textbf{82.6 $\pm$ 1.5} & \textbf{16.1 $\pm$ 4.2} & \textbf{0.5 $\pm$ 0.4} & 1.3 $\pm$ 0.7 \\ \hline
\end{tabular}}
\caption{Mean performances over all gender-dependent and gender-independent attributes on the validation set when using different methods to pick and label synthetic images. We find that most performances are comparable, with our method having a slightly higher AP, and slightly better DEO and KL.}
\label{tab:h_scores_incorrect}
\end{table}

\begin{comment}
\subsection{Gender expression manipulation}
\label{sec:gender_manip}
In this section, we provide some examples of images we generated by manipulating gender expression. Figure~\ref{fig:gender_exp_manipulation} shows three such examples. For each, we contrast our method with the method proposed by Denton et al.~\cite{DHMG19}. These images are chosen among the many generated in order to demonstrate the strength of our approach.  
\begin{figure}[ht!]
\centering
\scalebox{.5}{\input{gender_exp_manipulation}}
    \caption{Manipulation of gender expression from one gender expression through another by manipulating vectors in the latent vector space. The top two rows change from a masculine gender expression to a feminine gender expression, while the bottom 4 change from a feminine gender expression to a masculine one. For each attribute, the top row shows the naive manipulation, which preserves correlations of real-world data, and the bottom row shows our method. For example, for the attribute \texttt{Glasses}, we see that the naive manipulation removes eyeglasses while changing the gender expression from masculine to feminine, whereas our method preserves glasses.}
    \label{fig:gender_exp_manipulation}
\end{figure}
\end{comment}

\subsection{Number of required labeled images}
\label{sec:choi}
\begin{table}[t]
    \centering
    \resizebox{\linewidth}{!}{
    \begin{tabular}{|c|c|c|c|c|c|}
    \hline
    \multirow{2}{*}{Metric} & \multicolumn{5}{|c|}{Num. of samples used to compute $f_g$}  \\
    \cline{2-6}
    & 10 & 100 & 1000 & 10000 & 162,770 \\
    \hline         
    AP $\uparrow$ & 78.8 $\pm$ 1.5 & 78.8 $\pm$	1.5 &	78.8 $\pm$	1.5 & \textbf{78.9 $\pm$ 1.6} & 78.7 $\pm$ 1.6 \\ 
    DEO $\downarrow$ & 11.1 $\pm$ 3.4 & 11.3 $\pm$ 3.0 &	10.5 $\pm$ 3.7 & 10.8 $\pm$	3.7 & \textbf{9.6 $\pm$	3.1} \\ 
    BA $\downarrow$ & 0.6 $\pm$	0.5 & 1.0 $\pm$ 0.5 & 0.5 $\pm$ 0.5 & 0.7 $\pm$ 0.5 & \textbf{0.4 $\pm$ 0.5} \\
    KL $\downarrow$ & 0.6 $\pm$	0.2 & 0.8 $\pm$	0.3 & 0.7 $\pm$	0.3 & 0.7 $\pm$	0.3	& \textbf{0.5 $\pm$	0.6} \\ \hline
    \end{tabular}}
    \caption{Average over 4 attributes when using different numbers of labeled examples to compute gender expression. Results are reported on the validation set. We find that while the fairness metrics improve slightly by using more labelled examples, this is gradual, and within the error bars, in all cases.}
    \label{tab:choi_comp}
\end{table}

Choi et al.~\cite{GCSE19FairModeling} use a method that is unsupervised. Assuming access to a small unbiased dataset, as well as a large (possibly biased) dataset, they estimate the bias in the larger dataset, and learn a generative model that generates unbiased data at test time. Using these generated images, as well as real images, they train a downstream classifier for the attribute \texttt{Attractive}, and achieve an accuracy of 75\%. Since most of the protected attributes that we care about are sensitive (for example gender or race), not requiring protected attribute labels prevents perpetuation of harmful stereotypes. In order to understand how much our model depends on the protected attribute labels, we investigate where our model depends on the protected attributes labels. We use protected attribute labels only to compute the linear separator in the latent space ($\wg$ and $b_g$ from section~\ref{sec:derivation} in this document). We now train classifiers for gender expression, using different numbers of labeled images, and use these classifiers to train target attribute classifiers for 4 different attributes (\texttt{EyeBags}, \texttt{BrownHair}, \texttt{GrayHair} and \texttt{HighCheeks}).  
Most of the fairness metrics improve slightly when using more labeled examples (DEO improves from 11.1 when using just 10 samples to 9.6 when using all 162k samples in the CelebA training set, BA improves from 0.6 to 0.4, and KL improves from 0.6 to 0.5), however, these are all gradual, and within the error bars. Full results are in Table~\ref{tab:choi_comp}.

%{\small
%%\bibliographystyle{ieee_fullname}
%\bibliography{references}
%}

\end{document}